\documentclass{article}
\usepackage[ruled]{algorithm2e}
\usepackage{spconf,amsmath,graphicx}
\usepackage[colorlinks,citecolor=black]{hyperref}
\usepackage{bm}
\usepackage{amssymb}
\usepackage{booktabs}
\usepackage{multirow}
\usepackage{float}
\usepackage{graphicx} 
\usepackage{float} 
\usepackage{subfigure} 
\usepackage{graphbox}
\usepackage{graphics}
\usepackage{caption}
\usepackage{indentfirst} 

\title{Multi-view Contrastive Learning with Additive Margin for Adaptive Nasopharyngeal Carcinoma Radiotherapy Prediction}

\name{Jiabao Sheng$^{1,2}$, Yuanpeng Zhang$^{*3}$, Jing Cai$^{*1,2}$, \thanks{This work was supported in part by Shenzhen-Hong Kong-Macau S\&T Program (Category C) (SGDX20201103095002019), Shenzhen Basic Research Program (JCYJ20210324130209023) of Shenzhen Science and Technology Innovation Committee, Project of Strategic Importance (P0035421), Project of RISA (P0043001) of The Hong Kong Polytechnic University, the NSF of Jiangsu Province (No. BK20201441), Jiangsu Post-doctoral Research Funding Program (No. 2020Z020), and the NSFC (Grant No. 82072019).} Sai-Kit Lam$^{1,2}$, Zhe Li $^{1}$, Jiang Zhang$^{1}$, Xinzhi Teng$^{1}$}
\address{
$^{1}$ The Hong Kong Polytechnic University \\
$^{2}$ Research Institute for Smart Ageing, The Hong Kong Polytechnic University \\
$^{3}$ Department of Medical Informatics, Nantong University\\}

\begin{document}
%
\maketitle
\begin{abstract}
The prediction of adaptive radiation therapy (ART) prior to radiation therapy (RT) for nasopharyngeal carcinoma (NPC) patients is important to reduce toxicity and prolong the survival of patients. Currently, due to the complex tumor micro-environment, a single type of high-resolution image can provide only limited information. Meanwhile, the traditional softmax-based loss is insufficient for quantifying the discriminative power of a model. To overcome these challenges, we propose a supervised \textbf{m}ulti-view \textbf{con}trastive learning method with an additive \textbf{m}argin (MMCon). For each patient, four medical images are considered to form multi-view positive pairs, which can provide additional information and enhance the representation of medical images. In addition, the embedding space is learned by means of contrastive learning. NPC samples from the same patient or with similar labels will remain close in the embedding space, while NPC samples with different labels will be far apart. To improve the discriminative ability of the loss function, we incorporate a margin into the contrastive learning. Experimental result show this new learning objective can be used to find an embedding space that exhibits superior discrimination ability for NPC images.
\end{abstract}
\begin{keywords}
Medical Image Analysis, Multi-view, Nasopharyngeal Carcinoma, Contrastive Learning, Additive Margin
\end{keywords}

\section{Introduction}
\begin{figure}[ht] 
\includegraphics[width=0.5\textwidth]{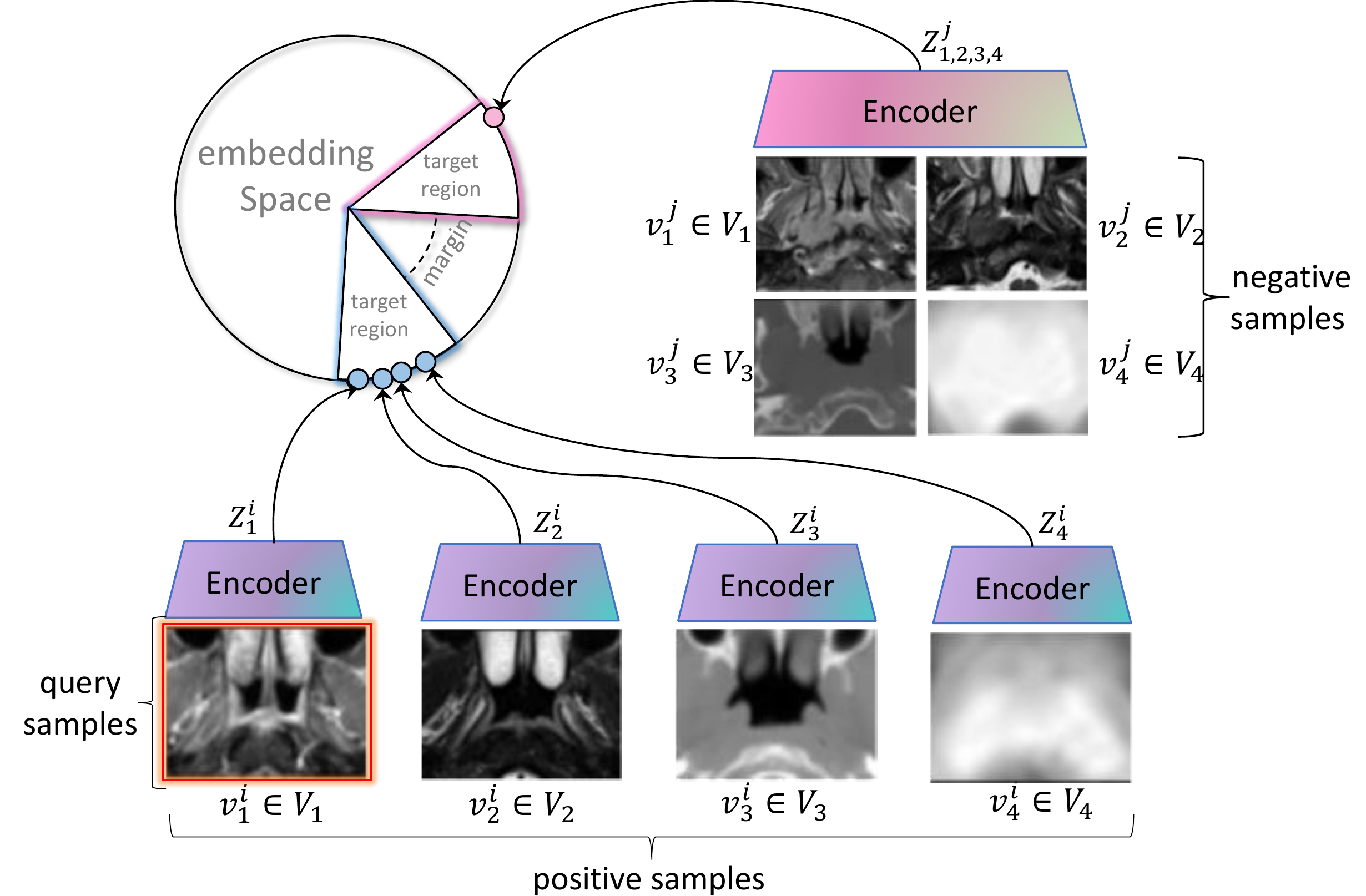} 
\caption{Illustration of our basic idea. $V_1$, $V_2$, $V_3$, and $V_4$ are different medical image views in the NPC-GTV dataset. $v^i$ and $v^j$ are different patients associated with NPC samples. $z^i$ and $z^j$ represent vectors. Our objective is to learn an embedding space in which similar sample pairs stay close while dissimilar ones are far apart.} 
\label{illustration} 
\end{figure}

Planning intensity-modulated radiotherapy (IMRT) for NPC requires medical imaging guidance. Previous studies have shown that the target volume (TV) and organ-at-risk (OAR) geometry appearing in images can change significantly during IMRT \cite{chung2014effects,lu2014assessment}. To reduce unnecessary exposure during treatment, it is necessary to incorporate medical image analysis to assist doctors in evaluating whether ART is needed.

In contrast to other medical image classification tasks, such as tumour identification \cite{saranya2021brain,sadad2021brain}, and cancer diagnosis \cite{tran2021deep,chen2021domain}, the prediction task for NPC ART is to analyze the properties of tumour to distinguish the need for radiotherapy replanning in the short term. Due to the heterogeneity of tumours \cite{el2021tumor, marusyk2010tumor}, the volume, shape, and texture of the tumour region may vary from patient to patient, and many diverse factors may cause these features to change.

In a previous study, \cite{ma2021mri} used artificially extracted magnetic resonance imaging (MRI) features to study radiation therapy planning for NPC. However, a single manually extracted omics signature cannot fully express the information of NPC samples \cite{lam2021multi}. The sample learning methods used in many studies  \cite{lam2021multi} showed that manually extracted multi-omics feature representation data could not be obtained better features than using effective deep learning algorithms. 
Leveraging better NPC feature representation to predict tumour variability remains a challenge for NPC ART.

With the resurgence of contrastive learning, significant advancements have recently been achieved in the learning of image representations \cite{wu2022distributed,zeng2021positional,yang2022proco,tian2020contrastive}. The research in \cite{tian2020contrastive} demonstrated how data augmentation based on various views of the same sample can benefit visual representation. However, \cite{dao2021multi} proved that the softmax-based contrastive learning loss is not directly applicable to image classification tasks, even though their work proposes a framework to improve the classification performance in contrastive learning, this method is not suitable for complex medical imaging tasks. 

To alleviate the above challenge, we propose \textbf{m}ultiview \textbf{m}argin \textbf{con}trastive (MMCon) learning, a medical image representation learning method. As shown in Fig~\ref{illustration}, given a set of NPC image views, a deep representation is learned by bringing views of the same class patient together in embedding space while pushing views of different class ones apart. We show an example of learned representation for 4-views(T1, T2, CT, and Dose). The embedding vector for each view may be concatenated to form the full representation of a patient.

The main contributions can be summarized as follows:
\begin{itemize}
\item We incorporated multi-view medical images to learn an NPC representation that aims to maximize the mutual information between different views of the same class patient by using positive pairs from various views.
\item We introduce a margin between distinct target class regions to achieve discriminative ability for unclear boundary samples via extending the conventional contrastive learning loss. With an extra margin, MMCon is more discriminative and noise-tolerant in the embedding space.
\end{itemize}

\section{Dataset Collection}
\label{sec:dataset}
We collected samples from 502 NPC patients who received radiotherapy in Hong Kong to construct the NPC-GTV dataset. Each patient has four different views, including CECT-T1w (T1 image), T2 MR (T2 image), CT images, and dose. All the planning images were retrospectively collected in the Digital Imaging and Communications in Medicine (DICOM) format and archived using an image archiving and communication system (PAC).  Patients who had clinical records regarded as necessitating the implementation of ART were labeled as 1; otherwise, patients were labeled as 0. The statistic of NPC-GTV is shown in Table \ref{statistic}.

The imaging data included planning CT images and pretreatment T1 and T2 MR images. The treatment-related data were the dose fractionation schemes. The outcome data included the replanning status and any replanning-related medical records. The attending radiation oncologists input all of the enrolled clinical records of the patients, which were carefully examined to determine the binary prediction outcome in this study. All CT and MR images were resampled to a voxel size of 1x1x1 mm$^3$ to mitigate the impacts of differences in image acquisition parameters among different patients. 
\begin{table}[ht]
\caption{The statistic of NPC-GTV dataset.}
    \centering
\resizebox{\linewidth}{!}{
\begin{tabular}{cccccc}
\toprule[2pt]
Organ &Views &Non-necessitating ART  &Necessitating ART &Samples &Total images\\ 
\hline
\multirow{4}{*}{GTVn}    &T1   &364  &138   &502 &\multirow{4}{*}{2,008}\\
                         &T2   &364  &138   &502 & \\
                         &CT   &364  &138   &502 & \\
                         &Dose &364  &138   &502 & \\
\toprule[2pt]
\end{tabular}
}
\label{statistic}
\end{table}

\section{Methodology}
\label{sec:methodology}
\begin{figure*} 
\centering
\includegraphics[width=\textwidth]{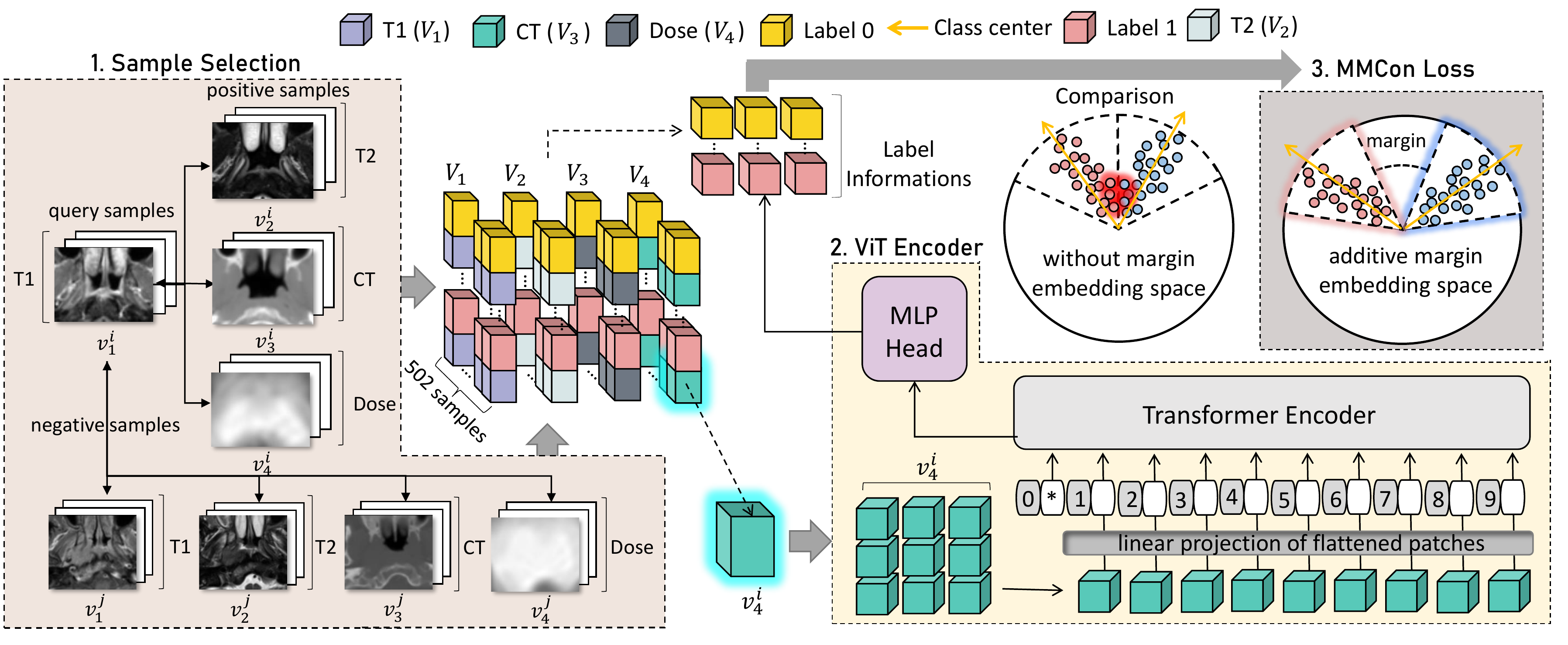} 
\caption{The whole framework of Multi-view Margin Contrastive Learning. We use NPC-GTV dataset to divide four different types of images (T1, T2, CT, Dose). 1. Sample Selection: The patient $i$ with $v_1^i$, $v_2^i$, $v_3^i$, $v_4^i$ can be constructed a serious of positive samples. Each view has 502 samples. Other samples from different patients $j$ belong to the negative samples. 2. ViT: Each sample of views will be encoded by ViT. 3. MMCon: The vectors from the encoder and the label information will be calculated in the MMCon loss function.} 
\label{structure} 
\end{figure*}

In this section, the representation learning framework in this study is first introduced. Then, this work proposes a marginal contrastive loss function with significant discrimination. Finally, we conclude by comparing the framework in this study and the contrastive loss function of the classification ability with multi-view to previous work. Our goal is to train a feature embedding network using labeled medical images. Embeddings for patient samples with similar diseases should be close to each other, while those from patients from different diseases should be far apart. The whole framework is shown in Fig.~\ref{structure}.

\subsection{Representation Learning Network}
Given a batch of input samples, we use different types of NPC images (T1, T2, CT, Dose) of the same organ to construct positive samples,  we regard them as multi-view medical images. The patient features of the embedding vector from the same instance should remain the same across various viewpoints, while the embeddings from different instances should be different. As shown in Figure ~\ref{illustration}, the multi-view instances are presented to the encoder network. At the output of the network, a margin contrastive loss is computed.

\textbf{Multi-view data} We match each input query sample to three different medical image views, each providing a unique view of the data. Among them, the T1 image is used as the query sample for each sample, and the remaining T2, CT, and Dose images from the same patient are used as positive samples relative to this sample. Other images from different patients belong to the negative sample. We set these views as $M$, which ${\cal M} = \{\bm{V}_1, \bm{V}_2, …,\bm{V}_m\}$. 

\textbf{Encoder Network} Our goal is to train an encoder network from a set of labeled images ${\cal X} = \{\bm{x}_1$, $\bm{f_{\theta}(\cdot) x}_2$, $ \ldots, \bm{x}_i\}$. $f_{\theta}(\cdot)$ converts the input image $\bm{x}_i$ to a low-dimensional embedding vector $\boldsymbol{h_i} = f_{\theta}\left(\bm{x} _{ i}\right) \in \mathbb{R}^d$, where $d$ is the output dimension. Both original and augmented samples are independently fed into the same type of encoder, resulting in four representation vectors. In this study, we chose ViT \cite{dosovitskiy2020image} as the encoder.
\subsection{Contrastive Loss Function}
\subsubsection{\textbf{Supervised Contrastive Losses}}
Supervised contrastive loss (SupCon)\cite{khosla2020supervised} can handle the situation where multiple samples are known to belong to the same class due to the presence of labels:
\begin{equation}
    L_{SupCon} = \sum_{i=1}^N \frac{-1}{\vert {\cal P}(i) \vert} \sum_{p \in {\cal P}(i)} \log \frac{\exp(\bm{z}_i \cdot \bm{z}_p /\tau)}{\sum_{a \in {\cal A}(i)}\exp(\bm{z}_i \cdot \bm{z}_a /\tau)} 
\label{Supcon}
\end{equation}
In Eq.~\ref{Supcon}, ${\cal P}(i)$ contains the indices of positive samples in the augmented batch (original + augmentation) with respect to $\bm{z}_i$ and $|{\cal P}(i)|$ is the cardinality of ${\cal P}(i)$. $\boldsymbol{z}_i$ is an anchor, it belongs to the query samples. $\boldsymbol{z}_a$ are negative samples. $\boldsymbol{z}_p$ are positive samples, and ${\cal A}(i)$ is the index set of negative samples.

\subsubsection{\textbf{Angular Margin based Contrastive Learning}}
In Eq.~\ref{Supcon}, the angular $\theta_{i,p}$ as follows:
\begin{equation}
    {\theta_{i,p}} = arccos(\frac{\bm{z}_i^\top,\bm{z}_p}{\lVert \bm{z}_i \rVert \cdot \lVert \bm{z}_p \rVert})
\label{theta}
\end{equation}

The decision boundary for $\bm{z}_{i} $ is $\theta_{i,p} = \theta_{i, a}$. A tiny perturbation around the decision boundary may result in an inaccurate conclusion if an insufficient decision margin is present.
To alleviate the problem, we proposed a new training objective for representation learning by adding an additive angular margin $m$ between positive pair $\bm{z}_i$, $\bm{z}_p$, and negative pair $\bm{z}_i$, $\bm{z}_a$, which can be formulated as follows:
\begin{equation}
\label{MarginCon}
\begin{aligned}
L&_{MarginCon} = \\ & \sum_{i=1}^{N} \frac{-1}{|P(i)|} \sum_{p \in P(i)} \log \frac{\exp(\cos(\theta_{i, p} + \alpha) / \tau)}{\sum_{a \in A(i)} \exp(\cos(\theta_{i, a} + \alpha) / \tau)}
\end{aligned}
\end{equation}

In this loss, $\alpha$ is the increased angle, the decision boundary for $\bm{z}_i$ is ${\theta_{i,p}+\alpha}={\theta_{i,a}}$. It increases the compactness of organ feature representation with the same semantics and enlarges the discrepancy of different semantic representations. This help enhances the alignment and uniformity properties, which are two key measures of representation quality related to contrastive learning, indicating how close between positive pair embeddings is and how well the embeddings are uniformly distributed. 

\subsubsection{\textbf{Multi-view Margin Contrastive Loss}}
Let multi-view sample as $M = \{V_1,V_2,…, V_m\}$. And divide them into three parts, which are query sample representation vector $\bm{z}_i$, positive samples representation vectors $\bm{z}_p$, and negative samples representation vectors $\bm{z}_a$. We bring samples from different views into the $L_{MarginCon}$, and ${\bm{z}_i} = {\bm{v}_1^i}$, ${z_p} = \{\bm{v}_2^i, … , \bm{v}_m^i\}$, ${\bm{z}_j}=\{\bm{v}_1^j, \bm{v}_2^j, …, \bm{v}_m^j \}$.
\begin{equation}
   sim(\bm{z}_i,\bm{z}_p) = \frac{f_{\theta 1}(\bm{v}_1^i) \cdot f_{\theta 2}(\{\bm{v}_2^i,…, \bm{v}_m^i\}}{\lVert f_{\theta 1}(\bm{v}_1^i) \rVert \cdot \lVert f_{\theta 2}(\{\bm{v}_2^i, … , \bm{v}_m^i\}) \rVert}
\label{CCL1}
\end{equation}
\begin{equation}
   sim(\bm{z}_i,\bm{z}_a) = \frac{f_{\theta 1}(\bm{v}_1^i)f_{\theta 2}(\{\bm{v}_1^j, … , \bm{v}_m^j\})}
   {\lVert f_{\theta 1}(\bm{v}_1^i) \rVert \ldotp \lVert f_{\theta 2}(\{\bm{v}_1^j, … , \bm{v}_m^j\}) \rVert}
\label{CCL2}
\end{equation}
Though the training objective tries to pull representations with similar images closer and push dissimilar ones away from each other, these representations may still not be sufficiently discriminative and not very robust to noise. By incorporating multi-view similarity and Eq.\ref{MarginCon}, We propose a \textbf{M}ulti-view \textbf{M}argin supervised \textbf{Con}trastive (MMCon) loss function for supervised embedding learning to improve the ability of decision classification as Eq.\ref{CCL4}.
\begin{equation}
\begin{aligned}
L&_{MMCon} = \\ & \sum_{i=1}^{N} \frac{-1}{|P(i)|} \sum_{p \in P(i)} \log \frac{\exp((sim(\bm{z}_i, \bm{z}_p)-m) / \tau)}{\sum_{a \in A(i)} \exp((sim(\bm{z}_i, \bm{z}_a)-m) / \tau)}
\end{aligned}
\label{CCL4}
\end{equation}
where $m$ is a margin, our margin is a scalar subtracted from $cos\theta$.


\section{EXPERIMENT}
\label{sec:experiment}
\subsection{Implementation Details}
We performed experiments on the NPC-GTV dataset. We use K-fold cross-validation for training and testing, where $k$ is 10. We set the margin $m$ to 0.2. The mini-batch size for training was 50. The contrastive learning temperature $\tau$ was set to 0.07. The learning rate was set to 0.001. Our experiments were run for 300 epochs. An SGD optimizer was used to optimize parameters. We used 3 A40 GPUs with 48G memory for training. This paper uses accuracy, precision, recall, and F1 value as metrics in binary classification. 

\subsection{Results and Analysis}
We evaluate the MMCon loss on the NPC-GTV dataset. We adopted image registration for the different view images which are CT image, T1 image, T2 image, and Dose. For the encoder network, we experimented with three different encoders \cite{he2016deep,dosovitskiy2020image,huang2017densely}, and three different loss which are contrastive learning loss \cite{khosla2020supervised}, Cross entropy loss and MMCon.

As shown in Table~\ref{result}, the performance of \cite{khosla2020supervised} is the worst among the three encoders. It divides the samples into three categories: query samples, positive samples, and negative samples. The query samples are compared with the positive and negative samples by learning the differences among them. However, the prediction task in this study is a fuzzy boundary classification problem. The results show that even after clustering, samples on the fuzzy boundary cannot be well classified.
Cross entropy achieves good performance by effectively leveraging the label information to ensure that samples of the same class are closely clustered. However, for NPC medical images, using only one view does not provide sufficient information for representation. Therefore, MMCon incorporates multi-view information and achieves best results than the other losses when used in combination with each encoder. MMCon also added a margin to the original contrastive learning loss function to ensure discriminative separation of the target and nontarget classes.
\begin{table}[ht]
\caption{The experiment of three different encoders and loss functions. For each encoder result, the best and second-best results in each metric are bold and underlined, respective.}
    \centering
\resizebox{\linewidth}{!}{
\begin{tabular}{crrrrr}
\toprule[2pt]
Encoder &Loss Function&Accuracy(\%) &Precision(\%) &Recall(\%)  &F1(\%)\\ 
\hline
\multirow{4}{*}{ResNet50 \cite{he2016deep}}&Supcon       &81.58&69.44&83.33&75.76\\
                         & Cross Entropy       &\underline{82.97}&\textbf{86.21}&\underline{86.21}&\underline{86.21}\\
                         &MMCon     &\textbf{90.67}&\underline{82.20}&\textbf{90.67}&\textbf{86.23}\\
\hline \hline
\multirow{4}{*}{DenseNet\cite{huang2017densely}}&Supcon        &77.28&70.95&75.86&72.58\\
                         & Cross Entropy        &\underline{80.79}&\textbf{85.94}&\underline{82.76}&\underline{84.32}\\
                         &MMCon     &\textbf{88.90}&\underline{83.06}&\textbf{91.14}&\textbf{86.91}\\
\hline \hline
\multirow{4}{*}{ViT \cite{dosovitskiy2020image}}    &Supcon        &80.85&75.77&79.31&77.14\\
                        & Cross Entropy      &\underline{86.90}&\underline{80.77}&\underline{89.87}&\underline{85.08}\\
                        &MMCon     &\textbf{91.28}&\textbf{83.42}&\textbf{91.33}&\textbf{87.20}\\
\toprule[2pt]
\end{tabular}
}
\label{result}
\end{table}

\subsection{Ablation Study}
As shown in Table~\ref{ablation}, we compare single-view images and multi-view images. In the single-view experiment, we used only 1 type of NPC image. Under the same loss function, three different encoders were used to conduct comparative experiments on T1 single-view, CT single-view, and T1+CT+T2+dose multi-view images. Due to space limitations, the experimental results obtained for T2 single-view and dose single-view images are omitted because they do not provide much useful analytical value. All three encoders show better results under the multi-view approach than under the single-view approach.

\begin{table}[ht]
\caption{The experiment of different views by using three different encoders with MMCon. For each encoder result, the best results in each metric are bold.}
    \centering
\resizebox{\linewidth}{!}{
\begin{tabular}{cccrrrr}
\toprule[2pt]
Loss & Encoder &Views &Accuracy(\%) &Precision(\%) &Recall(\%)  &F1(\%)\\ 
\hline
\multirow{3}{*}{MMCon} &\multirow{3}{*}{Resnet50\cite{he2016deep}} & T1 & 54.28 & 40.00 & 54.80 & 46.24\\
                                                      & & CT & 60.12 & 43.92 & 52.00 & 47.62\\
                                                      & & T1+T2+CT+Dose &\textbf{90.67} &\textbf{82.20} &\textbf{90.67} &\textbf{86.23}\\
                                                      \hline\hline
\multirow{3}{*}{MMCon}                           &\multirow{3}{*}{DenseNet\cite{huang2017densely}} &T1&52.70&40.32&55.17&46.59\\
                                                      & & CT&62.88&43.16&66.74&52.42\\
                                                      & &T1+T2+CT+Dose&\textbf{88.90}&\textbf{83.06}&\textbf{91.14}&\textbf{86.91}\\
                                                      \hline\hline
\multirow{3}{*}{MMCon}                           &\multirow{3}{*}{ViT\cite{dosovitskiy2020image}}      &T1&67.41&43.94&56.31&49.36\\
                                                      & & CT&64.20&50.37&58.29&54.04\\
                                                      & & T1+T2+CT+Dose&\textbf{91.28}&\textbf{83.42}&\textbf{91.33}&\textbf{87.20}\\
\toprule[2pt]
\end{tabular}
}
\label{ablation}
\end{table}

\section{Conclusion}
This study proposes a classification-capable supervised contrastive representation learning framework. We incorporate multi-view discrimination and an angular margin into the supervised contrastive learning loss to model the NPC image representation, thereby enhancing its discriminative ability. Experiments show that our architecture generally outperforms previous baselines on the NPC-GTV dataset.

\newpage
\bibliographystyle{IEEEbib}
\bibliography{refs}

\begin{thebibliography}{10}

\bibitem{chung2014effects}
Soon-Cheol Chung, Mi-Hyun Choi, Hyung-Sik Kim, Na-Rae You, Sang-Pyo Hong,
  Jung-Chul Lee, Sung-Jun Park, Ji-Hye Baek, Ul-Ho Jeong, Ji-Hye You, et~al.,
\newblock ``Effects of distraction task on driving: A functional magnetic
  resonance imaging study,''
\newblock {\em Bio-medical materials and engineering}, vol. 24, no. 6, pp.
  2971--2977, 2014.

\bibitem{lu2014assessment}
Jie Lu, Yidong Ma, Jinhu Chen, Liming Wang, Guifang Zhang, Mukun Zhao, and Yong
  Yin,
\newblock ``Assessment of anatomical and dosimetric changes by a deformable
  registration method during the course of intensity-modulated radiotherapy for
  nasopharyngeal carcinoma,''
\newblock {\em Journal of radiation research}, vol. 55, no. 1, pp. 97--104,
  2014.

\bibitem{saranya2021brain}
C~Saranya, J~Geetha Priya, P~Jayalakshmi, and E~Harini Pavithra,
\newblock ``Brain tumor identification using deep learning,''
\newblock {\em Materials Today: Proceedings}, 2021.

\bibitem{sadad2021brain}
Tariq Sadad, Amjad Rehman, Asim Munir, Tanzila Saba, Usman Tariq, Noor Ayesha,
  and Rashid Abbasi,
\newblock ``Brain tumor detection and multi-classification using advanced deep
  learning techniques,''
\newblock {\em Microscopy Research and Technique}, vol. 84, no. 6, pp.
  1296--1308, 2021.

\bibitem{tran2021deep}
Khoa~A Tran, Olga Kondrashova, Andrew Bradley, Elizabeth~D Williams, John~V
  Pearson, and Nicola Waddell,
\newblock ``Deep learning in cancer diagnosis, prognosis and treatment
  selection,''
\newblock {\em Genome Medicine}, vol. 13, no. 1, pp. 1--17, 2021.

\bibitem{chen2021domain}
Chen Chen, Yong Wang, Jianwei Niu, Xuefeng Liu, Qingfeng Li, and Xuantong Gong,
\newblock ``Domain knowledge powered deep learning for breast cancer diagnosis
  based on contrast-enhanced ultrasound videos,''
\newblock {\em IEEE Transactions on Medical Imaging}, vol. 40, no. 9, pp.
  2439--2451, 2021.

\bibitem{el2021tumor}
Nader El-Sayes, Alyssa Vito, and Karen Mossman,
\newblock ``Tumor heterogeneity: A great barrier in the age of cancer
  immunotherapy,''
\newblock {\em Cancers}, vol. 13, no. 4, pp. 806, 2021.

\bibitem{marusyk2010tumor}
Andriy Marusyk and Kornelia Polyak,
\newblock ``Tumor heterogeneity: causes and consequences,''
\newblock {\em Biochimica et Biophysica Acta (BBA)-Reviews on Cancer}, vol.
  1805, no. 1, pp. 105--117, 2010.

\bibitem{ma2021mri}
Xiangyu Ma, Xinyuan Chen, Jingwen Li, Yu~Wang, Kuo Men, and Jianrong Dai,
\newblock ``Mri-only radiotherapy planning for nasopharyngeal carcinoma using
  deep learning,''
\newblock {\em Frontiers in oncology}, vol. 11, pp. 713617, 2021.

\bibitem{lam2021multi}
Sai-Kit Lam, Yuanpeng Zhang, Jiang Zhang, Bing Li, Jia-Chen Sun, Carol Yee-Tung
  Liu, Pak-Hei Chou, Xinzhi Teng, Zong-Rui Ma, Rui-Yan Ni, et~al.,
\newblock ``Multi-organ omics-based prediction for adaptive radiation therapy
  eligibility in nasopharyngeal carcinoma patients undergoing concurrent
  chemoradiotherapy,''
\newblock {\em Frontiers in oncology}, vol. 11, 2021.

\bibitem{wu2022distributed}
Yawen Wu, Dewen Zeng, Zhepeng Wang, Yiyu Shi, and Jingtong Hu,
\newblock ``Distributed contrastive learning for medical image segmentation,''
\newblock {\em Medical Image Analysis}, vol. 81, pp. 102564, 2022.

\bibitem{zeng2021positional}
Dewen Zeng, Yawen Wu, Xinrong Hu, Xiaowei Xu, Haiyun Yuan, Meiping Huang, Jian
  Zhuang, Jingtong Hu, and Yiyu Shi,
\newblock ``Positional contrastive learning for volumetric medical image
  segmentation,''
\newblock in {\em International Conference on Medical Image Computing and
  Computer-Assisted Intervention}. Springer, 2021, pp. 221--230.

\bibitem{yang2022proco}
Zhixiong Yang, Junwen Pan, Yanzhan Yang, Xiaozhou Shi, Hong-Yu Zhou, Zhicheng
  Zhang, and Cheng Bian,
\newblock ``Proco: Prototype-aware contrastive learning for long-tailed medical
  image classification,''
\newblock in {\em International Conference on Medical Image Computing and
  Computer-Assisted Intervention}. Springer, 2022, pp. 173--182.

\bibitem{tian2020contrastive}
Yonglong Tian, Dilip Krishnan, and Phillip Isola,
\newblock ``Contrastive multiview coding,''
\newblock in {\em European conference on computer vision}. Springer, 2020, pp.
  776--794.

\bibitem{dao2021multi}
Son~D Dao, Ethan Zhao, Dinh Phung, and Jianfei Cai,
\newblock ``Multi-label image classification with contrastive learning,''
\newblock {\em arXiv preprint arXiv:2107.11626}, 2021.

\bibitem{dosovitskiy2020image}
Alexey Dosovitskiy, Lucas Beyer, Alexander Kolesnikov, Dirk Weissenborn,
  Xiaohua Zhai, Thomas Unterthiner, Mostafa Dehghani, Matthias Minderer, Georg
  Heigold, Sylvain Gelly, et~al.,
\newblock ``An image is worth 16x16 words: Transformers for image recognition
  at scale,''
\newblock {\em arXiv preprint arXiv:2010.11929}, 2020.

\bibitem{khosla2020supervised}
Prannay Khosla, Piotr Teterwak, Chen Wang, Aaron Sarna, Yonglong Tian, Phillip
  Isola, Aaron Maschinot, Ce~Liu, and Dilip Krishnan,
\newblock ``Supervised contrastive learning,''
\newblock {\em Advances in Neural Information Processing Systems}, vol. 33, pp.
  18661--18673, 2020.

\bibitem{he2016deep}
Kaiming He, Xiangyu Zhang, Shaoqing Ren, and Jian Sun,
\newblock ``Deep residual learning for image recognition,''
\newblock in {\em Proceedings of the IEEE conference on computer vision and
  pattern recognition}, 2016, pp. 770--778.

\bibitem{huang2017densely}
Gao Huang, Zhuang Liu, Laurens Van Der~Maaten, and Kilian~Q Weinberger,
\newblock ``Densely connected convolutional networks,''
\newblock in {\em Proceedings of the IEEE conference on computer vision and
  pattern recognition}, 2017, pp. 4700--4708.

\end{thebibliography}

\end{document}